\documentclass[runningheads]{llncs}
 
\usepackage{eccv}

\usepackage{eccvabbrv}

\usepackage{graphicx}
\usepackage{booktabs}
\usepackage[accsupp]{axessibility}  
\usepackage{caption}
\usepackage{subcaption}
\usepackage{makecell}
\usepackage{wrapfig}
\usepackage{arydshln}
\newcolumntype{Y}{>{\centering\arraybackslash}X}

\usepackage[pagebackref,breaklinks,colorlinks,citecolor=eccvblue]{hyperref}

\begin{document}

\title{Task-Specific Adaptation of Segmentation Foundation Model via Prompt Learning}

\titlerunning{Task-specific adaptation of segmentation
foundation model}

\author{Hyung-Il Kim\thanks{Both authors equally contributed.}\inst{1}\and
Kimin Yun$^\star$\inst{1}\and
Jun-Seok Yun\thanks{This work was done during his internship at ETRI. }\inst{2}\and Yuseok Bae\inst{1}}

\authorrunning{H.-I. Kim and K. Yun et al.}

\institute{ETRI, Daejeon, Republic of Korea\\ \email{\{hikim, kimin.yun, baeys\}@etri.re.kr}\and
KITECH, Daegu, Republic of Korea
\\ \email{yunjs@kitech.re.kr}}

\maketitle
\vspace{-0.2cm}
\begin{abstract}
Recently, foundation models trained on massive datasets to adapt to a wide range of tasks have attracted considerable attention and are actively being explored within the computer vision community. 
Among these, the Segment Anything Model~(SAM) stands out for its remarkable progress in generalizability and flexibility for image segmentation tasks, achieved through prompt-based object mask generation. 
However, despite its strength, SAM faces two key limitations when applied to instance segmentation that segments specific objects or those in unique environments (\eg, task-specific adaptation for out-of-distribution objects) not typically present in the training data: 1) the ambiguity inherent in input prompts and 2) the necessity for extensive additional training to achieve optimal segmentation. 
To address these challenges, we propose a task-specific adaptation (\ie, customization) of the segmentation foundation model via prompt learning tailored to SAM.
Our method involves a prompt learning module (PLM), which adjusts input prompts into the embedding space to better align with peculiarities of the target task, thereby enabling more efficient training. Furthermore, we introduce a point matching module (PMM) to enhance the feature representation for finer segmentation by ensuring detailed alignment with ground truth boundaries. 
Experimental results on various customized segmentation scenarios demonstrate the effectiveness of the proposed method. 
\end{abstract}

\section{Introduction}
\label{sec:intro} 
To identify and segment pixels belonging to each object instance, instance segmentation technology has been considered a crucial component for high-level scene understanding. 
In addition to general instance segmentation trained with common object instances~(\eg, COCO dataset~\cite{coco}), instance segmentation for segmenting specific objects~(\eg, face, salient object) has been widely studied for various real-world applications: autonomous driving~\cite{segment_for_auto_driving_IEEE_TITS2020, segment_for_auto_driving_CVPRW2018}, medical image segmentation~\cite{C-cam_CVPR2022, medical_seg_CVPR2023, Cyclemix_CVPR2022}, and image editing~\cite{Imag_editing_SPL2020, Editgan_Neurips2021}. 

\begin{figure}[t]
\centering
\includegraphics[height=4.0cm]{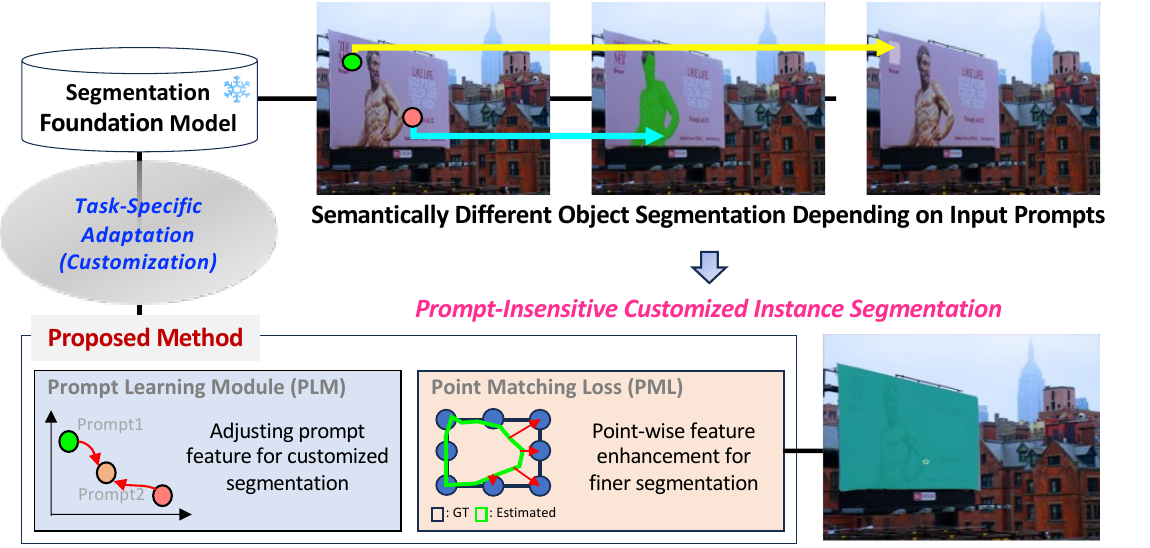}
\vspace{-0.1cm}
\caption{Our proposed method mitigates SAM's sensitivity to input prompts by adjusting prompt features in the embedding space to align with class-wise object mask-based user intentions via a prompt learning module~(PLM). Additionally, we enhance the feature representation for finer object segmentation through training with a point matching module (PMM).}
\vspace{-0.6cm}
\label{fig:teaser}
\end{figure}
 
Recent advances in deep learning have led to significant progress in instance segmentation algorithms.
Inspired by the Faster R-CNN~\cite{ren2015faster}, two-stage object detector-based instance segmentation algorithms~(\eg, Mask R-CNN~\cite{maskrcnn} and Mask Scoring R-CNN~\cite{huang2019mask}) have been introduced, followed by one-stage object detector-based methods~(\eg, YOLACT~\cite{yolact}). 
To address the limitations of relying on pre-defined anchors, anchor-free algorithms~\cite{solo, centermask} have been introduced.
Furthermore, the integration of attention mechanisms from Transformer architectures~\cite{attention, vit} has led to the development of query-based instance segmentation methods, such as ISTR~\cite{istr}, QueryInst~\cite{queryinst}, and SOLQ~\cite{solq}.
Despite these technological advances, instance segmentation still faces significant challenges, particularly in adapting to a variety of environments and improving training efficiency while minimizing the reliance on costly mask-annotated data.

More recently, foundation models~\cite{foundation}, trained with vast datasets to adapt to a wide range of downstream tasks, have received tremendous attention in the field of computer vision~\cite{clip, align, dino}. 
Among them, a segmentation foundation model known as the `Segment Anything Model~(SAM)'~\cite{sam} has received the spotlight regarding generalizability and flexibility in image segmentation. 
Trained on vast datasets, SAM is designed as a promptable model, adept at generating segmentation masks for object regions in response to user prompts, such as points or text.
Despite its powerful generalization capability and flexibility, the SAM still faces challenges in segmenting specific objects or those in unique environments not covered in its training data. 
As depicted in Fig.~\ref{fig:teaser}, SAM's sensitivity to input prompts causes a significant issue. 
The inherent ambiguity in these prompts can result in substantial variations, affecting both the segmented object type and the segmentation mask quality. 
Moreover, optimizing the segmentation model for uniquely shaped objects in particular environments (\eg out-of-distribution (OOD) objects) requires extensive training with additional large-scale datasets.

To tackle these challenges, we focus on task-specific adaptation (\ie, customization) of the SAM to better reflect user intention in instance segmentation, especially when users collect datasets for specific segmentation targets.
SAM is a promptable model responsive to user intention, as illustrated in Fig.~\ref{fig:teaser}. Its sensitivity to input prompts can often lead to repetitive attempts or even failures in precisely segmenting the desired objects. To address this, our approach provides a customized segmentation for task-specific adaptation of the SAM with an additional learning module, utilizing datasets collected by users with mask annotations for the specific objects the user aims to segment.

Specifically, to mitigate the issue of sensitivity to input prompts, we devise a prompt learning module~(PLM). This module transforms input prompts within the embedding space to accurately reflect the user's intentions for customized segmentation. A key advantage of the PLM is its plug-and-play capability, which allows for efficient customization of the segmentation model. By selectively training only the PLM while keeping the rest of the model weights frozen, it enables effective adaptation without extensive training.
Additionally, we introduce a point matching module~(PMM) to enhance the segmentation model's performance further. PMM improves feature representation for finer segmentation by focusing on features to object boundary points. 
We validate the efficacy of our proposed method through experiments focused on customized instance segmentation tasks: facial part segmentation, outdoor banner segmentation, and license plate segmentation. Our findings consistently demonstrate the effectiveness of our approach in instance segmentation tasks tailored to user intention. 

The major contributions of our work are as follows: 
\begin{itemize}
    \item Our method effectively tackles the problem of prompt sensitivity in the SAM, leading to more robust instance segmentation that adheres to user-intended object shapes. This approach ensures the segmentation process is not only accurate but also aligned with specific user requirements.
    \item Our approach leverages a plug-and-play prompt learning module, enabling efficient task-specific adaptation without comprehensive fine-tuning. Notably, this method preserves the foundational model's generalizability, making it a versatile solution across diverse segmentation tasks.
    \item Furthermore, a point matching module is devised to enhance features for boundary points on a mask, which contributes to finer segmentation. 
\end{itemize}

\section{Related Work} 
\label{sec:related}
\subsection{Segment Anything}
As a foundation model in computer vision, SAM~\cite{sam} has recently shown remarkable zero-shot image segmentation performance by harnessing the power of a large-scale dataset containing over $1B$ mask data. 
This model's outstanding generalization capability has the potential to be applied to image understanding tasks across diverse environments~\cite{med_SAM, med_SAM_ada, SAM-OCTA, SAM_Nerfs, SAM_remote_sensing, Track_anything, SAM_Track, Inpaint_anything, cao2023segment}. 
For instance, MedSAM~\cite{med_SAM} adapted SAM for medical image segmentation using $1M$ medical image-mask pairs. 
In the 3D image understanding, Cen \textit{et al.}~\cite{SAM_Nerfs} proposed 3D object segmentation through cross-view self-prompting and mask inverse rendering, utilizing single-view 2D masks generated by SAM. 
The tracking anything module~(TAM)~\cite{Track_anything} was designed to assess and refine the quality of SAM-initiated masks for video object segmentation, aiming to address SAM's inconsistencies in mask estimation across video frames.
In addition to the task-specific approaches, there have been efforts to analyze and improve SAM~\cite{Mazurowski_et_al, zhang_et_al, Tang_et_al, Zhou_et_al}. 
SAM-OCTA~\cite{SAM-OCTA} fine-tuned the SAM encoder's parameters using the low-rank adaptation~\cite{LoRA} to adapt specific datasets while preserving the semantic understanding of SAM. The medical SAM adapter~\cite{med_SAM_ada} was proposed to integrate medical-specific domain knowledge into the SAM using a simple adapter. 

While SAM's effectiveness has been recognized, fully exploiting the foundation model's potential requires further exploration in additional training and prompting strategies. 
In this context, we address the SAM's sensitivity to input prompts and the task-specific adaptation (\ie, customization) of the foundation model for specific object segmentation. 
To this end, we propose a prompt learning-based task-specific adaptation method, enabling specific object segmentation while building upon SAM's generalization capabilities. 

\subsection{Prompt Tuning}
Foundation models have been utilized for specific downstream tasks through retraining or fine-tuning. 
These methods typically demand considerable computational resources, large-scale datasets, and significant time.
To address these issues, the concept of prompt tuning~\cite{prompt_NLP, Prefix-tuning, Power_prompt_tuning, P-tuning_v2} has received attention, particularly in natural language processing~(NLP).
This approach utilizes the inherent knowledge of foundation models without necessitating the retraining of the entire model.
Inspired by these studies in NLP, visual prompt tuning~(VPT)~\cite{VPT} has demonstrated remarkable adaptation performance in computer vision by training minimal prompt parameters. 
VPT has been shown to be superior to other adaptation approaches, such as full model training or head-oriented tuning, in terms of both effectiveness and efficiency. 
Recently, prompt tuning approaches based on SAM have been actively explored. 
PerSAM~\cite{PerSAM} has proposed to personalize segmentation, aiming to identify areas in images that share the same foreground as a user-provided image.
It introduces additional inputs to SAM's decoder, such as target-guided attention and prompting mechanisms, thereby enabling personalized segmentation.
HQ-SAM~\cite{HQ-SAM} has tackled the degradation of mask quality in complex structures by using a learnable output token, enhancing mask details through the aggregation of global-local features and fine-tuning small parameters with fine-grained mask datasets.

In this regard, we propose a prompt learning method designed to segment specific object instances by customizing the segmentation foundation model. 
Our approach freezes the SAM's model parameters, focusing on training only the proposed prompt learning module for task-specific adaptation.
The proposed method dynamically modulates the prompt features depending on input for task customization in embedding space, by learning the shape prior for segmentation.
This approach not only streamlines the process but also facilitates efficient, plug-and-play task-specific adaptation for customized segmentation tasks.


\begin{figure}
    \centering
    \begin{subfigure}[]{0.46\textwidth}
        \centering
        \includegraphics[width=\textwidth]{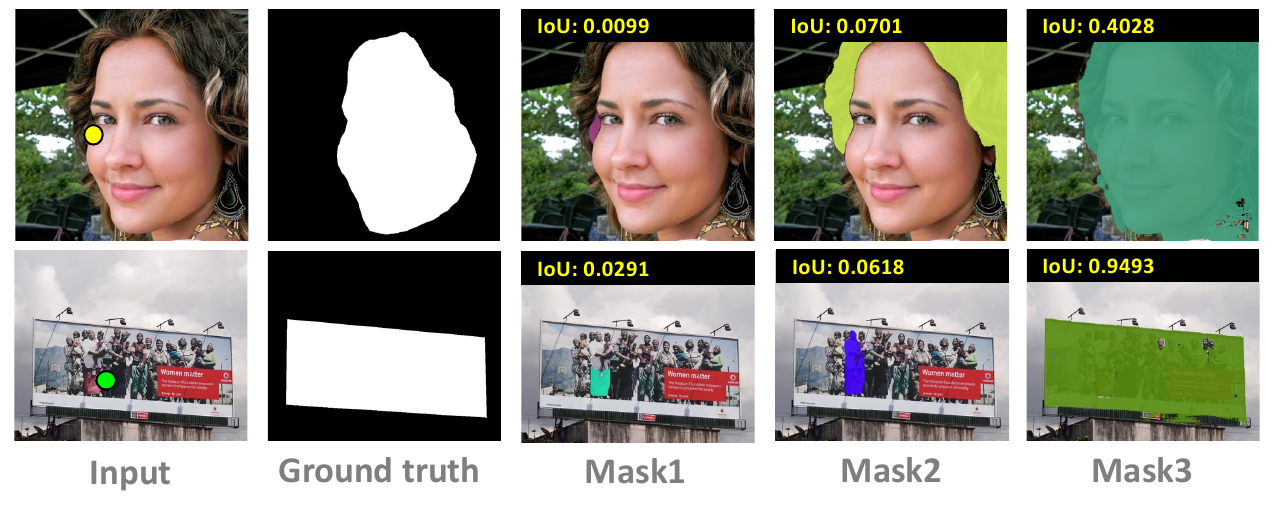}
        \caption{}
        \label{fig:sam}
    \end{subfigure}\hfill
    \begin{subfigure}[]{0.465\textwidth}
        \centering
        \includegraphics[width=\textwidth]{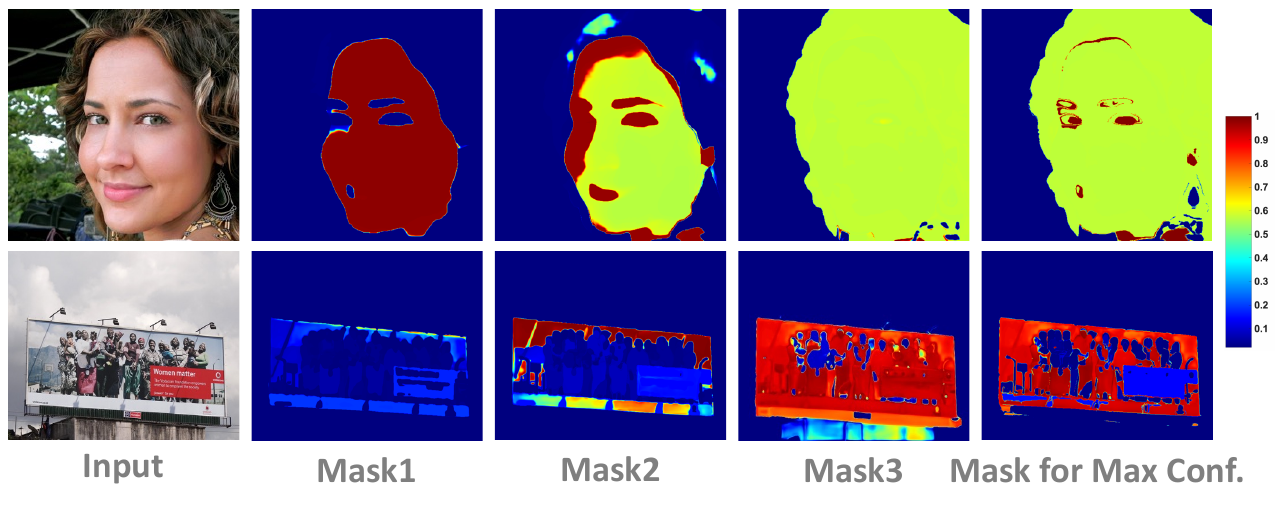}
        \caption{}
        \label{fig:errmap}
    \end{subfigure}
    \vspace{-0.3cm}
    \caption{(a) Instance segmentation results with the SAM for ambiguous input prompts and (b) visualization of IoU maps for multiple masks estimated by the SAM, where each pixel denotes the IoU value between the GT mask and the estimated mask. Note that each pixel location means the location of the input prompt.}
    \label{fig:qual_face1}
    \vspace{-0.3cm}
\end{figure}

\section{Ambiguity in Prompts} 
\label{sec:ambiguity}
In this section, we investigate the prompt ambiguity problem in SAM, one of the foundation models targeted in this paper. SAM's architecture comprises an image encoder, a prompt encoder, and a mask decoder. 
Given the input image~($x$), SAM estimates the segmentation mask~($m$) of the object according to the user’s input prompt~($p$) as follows: 
\begin{equation}
    m=D\left(E_{I}(x), E_{P}(p)\right), 
\end{equation}
where $E_{I}$, $E_{P}$, and $D$ denote the image encoder, prompt encoder, and mask decoder, respectively. 
As pointed out in~\cite{sam}, when $p$ is given ambiguously~(\eg, located at the boundary of an object or in an area where two different objects overlap), semantically different objects can be segmented. 
To deal with this problem, the approach in~\cite{sam} 
generates three output masks based on the confidence score~(\ie, the estimated intersection over union (IoU)), using a small number of output tokens. 
This model was trained through the backpropagation of the minimum loss between the ground truth (GT) and the three estimated masks.

Despite SAM's flexibility to infer multiple masks, there are still issues to consider when employing the segmentation foundation model in customized instance segmentation scenarios. 
To investigate the prompt ambiguity problem, let us consider two segmentation scenarios: 1) face segmentation, where the user aims to segment only the skin region, and 2) outdoor banner segmentation, focusing on segmenting the whole banner area.
As illustrated in Fig.~\ref{fig:sam}, if the prompt is near the face's border, the segmented area may include unintended parts, like hair.
Even in the mask results about the whole face among multiple masks, we observe that the face including the hair is segmented, rather than the skin desired by the user. 
Similarly, in outdoor banner segmentation, depending on the prompt's placement, unintended objects like parts of a person can be segmented, deviating from the user’s intention.
To further examine the sensitivity of input prompts, we visualize IoU values comparing the GT mask with the SAM output mask by moving the prompt to all positions in Fig.~\ref{fig:errmap}, where red indicates an IoU value close to $1$, and blue indicates a value near $0$.
As expected, it shows that successful segmentation can be made when the input prompt exists in a place where there are no specific object instances that differ from the intention~(\eg, the unobstructed skin area of the face or the flat banner area without
\vspace{-0.4cm}
\begin{figure*}
\centering
\includegraphics[height=2.25cm]{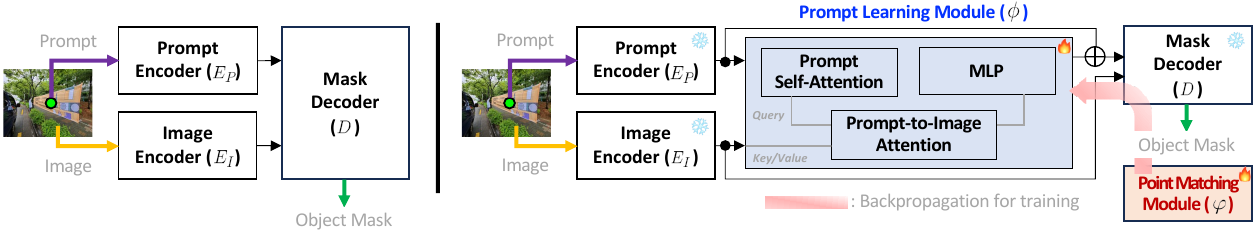}
\vspace{-0.3cm}
\caption{Overall framework of the proposed method. Building upon the SAM~(left) with two encoders and a mask decoder, the proposed method~(right) introduces two additional modules. The prompt learning module~(PLM) $\phi$ adjusts the prompt feature so that the user's desired object can be segmented well. In addition, the point matching module~(PMM) $\varphi$ enables finer segmentation through learning to minimize the distance between the GT points and estimated points by $\varphi$.}
\label{fig:proposed}
\vspace{-0.6cm}
\end{figure*}
people or text). 
This is mainly because the SAM predicts the most probable contour containing the prompt, rather than aligning with a specific user intention.
Consequently, users need to make fine adjustments to their input prompts to achieve effective segmentation. To streamline this process, we introduce a task-specific adaptation (\ie, customization) instance segmentation approach that refines prompts for more accurate segmentation of the user's intended objects by leveraging the generalization capabilities of SAM.



\section{Proposed Method} 
\label{sec:proposed}
In this paper, we propose a task-specific adaptation of a segmentation foundation model, \ie, SAM, for customized instance segmentation through prompt learning. 
To address the sensitivity issue to input prompts, we devise a prompt learning module~(PLM).
This module transforms input prompts within the embedding space, reflecting the user's intention for customized segmentation. 
Additionally, we introduce a point matching module~(PMM) to enhance segmentation performance by focusing on features related to object boundary points, which aids in matching the segmentation more closely to the ground truth boundary.

As shown in Fig.~\ref{fig:proposed}, given an input image~($x_{i}$) and a prompt~($p_{i}$), image and prompt features~($f_{I}^{i}$ and $f_{P}^{i}$) are extracted by the image and prompt encoders~($E_{I}$ and $E_{P}$), \ie, $f_{I}^{i} = E_I\left(x_{i}\right)$ and $f_{P}^{i} = E_P\left(p_{i}\right)$. 
As discussed in Section~\ref{sec:ambiguity}, the PLM is designed to address the issue of ambiguous input prompts by learning to transform the $f_{P}^{i}$.
The training process of PLM is driven by a dataset containing instances of the user's target objects, teaching the PLM how to modify the $f_{P}^{i}$ effectively. 
It does this by estimating necessary adjustments within the embedding space, which are informed by both $f_{I}^{i}$ and $f_{P}^{i}$.
Through this training, the PLM learns the optimal transformation of the prompt feature as a residual form, aligning it more closely with the specific segmentation needs of the user.
Next, based on this transformed prompt feature and original image embedding, the object instance mask $m_{i}$ is obtained by a mask decoder $D$. 
Note that the PMM is introduced in the training phase to improve the quality of the mask generated by $D$. 
By focusing on features related to boundary points extracted by the mask decoder, the PMM improves the precision of the segmentation contours.
To maintain the generalization capability of the segmentation foundation model, we keep the architecture of the image encoder, prompt encoder, and mask decoder unchanged, freezing their pre-trained weights.

\subsection{Prompt Learning Module}
The PLM $\phi$ aims to adjust prompts, ensuring effective segmentation of the desired object by the user, regardless of the initial prompt provided for instance segmentation. 
Our approach primarily utilizes sparse prompts~(\eg, point or bounding box) that the users can easily provide.
Instead of making adjustments in a low-dimensional space~($p_{i}\in\mathbb{R}^{2}$) with limited information, the PLM operates in a higher-dimensional space~($f_{P}^{i}\in\mathbb{R}^{256}$) for prompt feature adjustments.

The PLM, depicted in Fig.~\ref{fig:proposed}, consists of three consecutive operations: self-attention of prompt features, prompt-to-image attention~(with the attended prompt features as queries), and multi-layer perception~(MLP).
The PLM, utilizing multi-head~($T$) attention, calculates the necessary adjustment for the prompts in the embedding space based on the image features $f_{I}^{i}$ and prompt features $f_{P}^{i}$. 
This adjustment (offset) is represented as $\bigtriangleup f_{P}^{i}$. 
Specifically, in the self-attention block, the $f_{P}^{i}$ is embedded based on the attention operation between tokens in the $f_{P}^{i}$. 
Next, in the prompt-to-image attention block, an attention operation is performed using the prompt features embedded by the self-attention as the query and the $f_{I}^{i}$ as the key/value. 
Then, the MLP layer updates each token of the attended embedding. 
Each attention block includes positional embedding to its inputs, and a layer normalization~\cite{layernormalization} operation follows each block.
In summary, the PLM estimates the prompt feature change offset $\triangle f_{I}^{i}$, given the $f_{I}^{i}$ and $f_{P}^{i}$, to enhance a segmentation result: 
\begin{equation}
    \triangle f_{P}^{i} =  \phi\left(f_{P}^{i}, f_{I}^{i}\right). 
\end{equation}
Then, the transformed prompt feature $\tilde{f}_{P}^{i}$ is obtained by adding the estimated offset $\triangle f_{P}^{i}$ to the $f_{P}^{i}$, \ie, $\tilde{f}_{P}^{i} = f_{P}^{i} + \triangle f_{P}^{i}$. 
Based on $\tilde{f}_{P}^{i}$ and $f_{I}^{i}$, the segmentation mask $\tilde{m}_{i}$ is then estimated by the mask decoder, $\tilde{m}_{i}=D(f_{I}^{i}, \tilde{f}_{P}^{i})$.  
That is, the ambiguity problem is handled by learning to estimate the optimal prompt embedding change $\triangle f_{P}^{i}$ from random prompt samples in training.

\subsection{Point Matching Module}
To enhance the mask decoder’s ability to estimate the segmentation mask $\tilde{m}_{i}$ using the adjusted prompt feature, we introduce the PMM $\varphi$, during training, focusing on refining features, particularly those associated with mask details and quality. 
Motivated by TextBPN++~\cite{textbpn++}, the proposed PMM is designed to guide mask edge points toward the corresponding GT points.
In other words, this involves learning the offset required to align with these GT points as an auxiliary task.
Unlike TextBPN++, which updates points iteratively, our module estimates these points through an end-to-end training approach. 
However, since extracting boundary points from the mask via contour fitting is non-differentiable, this process cannot be trained through standard backpropagation. 
To deal with this issue, we utilize points on the mask edge of the GT mask, augmented with jittering for training purposes.
Suppose that the set of the GT points and its jittered points are $\mathcal{G}_{i}=\left\{c_{i}^{1},c_{i}^{2},\cdots,c_{i}^{K} \right\}$ and $\mathcal{G}_{i}^{\ast}=\left\{c_{i}^{\ast 1}, c_{i}^{\ast 2},\cdots,c_{i}^{\ast K} \right\}$, respectively. 
And the feature extracted by the two-way attention block in the mask decoder, denoted as $f_{D_{T}}^{i}$, is used for extracting point features. 
However, since the feature map size $f_{D_{T}}^{i}\in\mathbb{R}^{H_{f}\times W_{f}\times C}$ differs from the jittered points' coordinates $\mathcal{G}_{i}^{\ast}$ in image space $x_{i}\in\mathbb{R}^{H_{I}\times W_{I}}$ ($H_{I}> H_{f},\, W_{I}> W_{f}$), 
we align the coordinate of $\mathcal{G}_{i}^{\ast}$ to the feature map space through interpolation.
This step allows us to collect a feature matrix $\mathcal{W}_{i}\in\mathbb{R}^{K\times C}$ for each point in $\mathcal{G}_{i}^{\ast}$.
For convenience, we define this function (\ie, interpolation and point feature extraction) as $c(\cdot)$, \ie, $\mathcal{W}_{i} = c(f_{D_{T}}^{i}, \mathcal{G}_{i}^{\ast})$: 
Based on the $\mathcal{W}_{i}$ for $\mathcal{G}_{i}^{\ast}$, the boundary transformer~\cite{textbpn++} with the encoder-decoder structure is utilized to estimate the refined boundary points $\mathcal{\tilde{G}}_{i}$. 
Following the design in~\cite{textbpn++}, the encoder consists of three transformer blocks with residual connections, and the decoder employs a three-layered $1\times1$ convolution with ReLU activation. 
In summary, the PMM $\varphi$ refines $\mathcal{G}_{i}^{\ast}$ from $\mathcal{W}_{i}$, thus returning $\mathcal{\tilde{G}}_{i} = \varphi(\mathcal{W}_{i})$. The set $\tilde{\mathcal{G}}_{i}$ of the refined boundary points is used for training the proposed network. 

\subsection{Training}
Our network integrates two proposed modules with the SAM for end-to-end training.
In this setup, all parameters within the SAM are frozen.
The objective function $\mathcal{L}$ for training consists of the sum of two loss functions like:  
\begin{equation}
    \mathcal{L}=\frac{1}{N}\sum_{i=1}^{N}\left(\mathcal{L}_{seg}(m_i,\tilde{m}_{i}) + \lambda\mathcal{L}_{pm}(\mathcal{G}_{i}, \tilde{\mathcal{G}}_{i})\right),
    \label{eq:loss}
\end{equation}
where $\mathcal{L}_{seg}$ and $\mathcal{L}_{pm}$ denote the loss functions for segmentation and point matching, respectively. And, $\lambda$ is a balancing hyperparameter. 
Regarding $\mathcal{L}_{seg}$, we follow the loss function detailed in~\cite{sam}, which computes a linear combination of focal loss and dice loss at a 20:1 ratio for the GT mask $m_{i}$ and the estimated mask $\tilde{m}_{i}$. Besides, IoU loss and mask loss to predict the IoU score itself are used to compute the $\mathcal{L}_{seg}$.  
The loss function $\mathcal{L}_{pm}$ is designed to enhance the precision of mask estimation through the PMM $\varphi$, which is defined as 
\begin{equation}
    \mathcal{L}_{pm}(\mathcal{G}_{i},\tilde{\mathcal{G}}_{i}) = \frac{1}{K}\sum_{k=1}^{K} \left(\underset{\tilde{c}_{i}^{j}\in\tilde{\mathcal{G}_{i}}}{\mathrm{inf}}\,\{\|c_{i}^{k}-\tilde{c}_{i}^{j} \|^{2}\} \right). 
\end{equation}
This function computes the distance between the set ($\mathcal{G}_{i}$) of points for the GT and the points ($\tilde{\mathcal{G}}_{i}$) predicted by the $\varphi$. 

\section{Experiments}
\label{sec:experiments}

\subsection{Experimental Settings}
In this section, we conform to the single-point valid mask evaluation method as established in the original SAM study. Our analysis spans three tasks: facial part segmentation, outdoor banner segmentation, and license plate segmentation. The evaluation focuses on segmenting an object from a single foreground point. Given that SAM can predict three masks, we assess both approaches: the model’s most confident mask and scenarios involving multiple masks. In cases of multiple masks, we report the highest IoU, assuming hypothetical user selection from the three masks. We refer to this as SAM (oracle), consistent with the terminology in~\cite{sam}. We employ the standard mean IoU (mIoU) metric to evaluate the match between predicted and GT masks. 
We report results in four scenarios: standard results for SAM+PLM and SAM+PLM+PMM, plus their oracle versions (SAM+PLM~(oracle) and SAM+PLM+PMM~(oracle)), showcasing the effectiveness of our method under mask selection conditions as in~\cite{sam}.

In a training phase, to align with how users typically interact with the segmentation model, often choosing plausible foreground positions over exact centers, we refined our training approach. An arbitrary point from the training mask was selected as the input prompt, with the corresponding GT mask as the target output. 
We employed a probabilistic function to determine prompt positions, prioritizing those close to the center.
%
During the testing phase, we evaluated the model's accuracy using the center point of the mask as the input. 
For comparison, we implemented a method named SAM-F by applying the scale-aware fine-tuning proposed in PerSAM~\cite{PerSAM}. This approach, adapted from the SAM, utilizes three scales of masks, \(M_1\), \(M_2\), and \(M_3\), and employs two learnable parameters, \(w_1\) and \(w_2\), to finetune the final mask such that \(M = w_1M_1 + w_2M_2 + (1-w_1-w_2)M_3\). Through this, we aim to evaluate the capability to achieve a user-desired mask by only combining the output scales of SAM.

Our model builds upon the pre-trained SAM, \ie, MAE pre-trained ViT-H image encoder. Our updates are confined to the PLM (1.6M parameters) and the PMM (1.2M parameters) while keeping the other parameters (641M of SAM-ViT-H) frozen. In testing, only the parameters of the PLM $\phi$ (except $\varphi$) are utilized. However, the parameters for $\varphi$ can also be employed in certain scenarios through a two-step process (please refer to the discussion of Fig.~\ref{fig:qual_refine}). 
For our model optimization, we set the learning rate to $0.00001$ and utilized an AdamW optimizer with a weight decay of $0.01$. The model was trained on eight A40 GPUs over $20,000$ steps with a batch size of $10$ and an image size of $1,024\times1,024$ pixels. A learning rate scheduler incorporating a warmup phase incrementally increased the learning rate for the initial $250$ steps. After the warmup period, we applied a stepped decay approach to the learning rate, reducing at predetermined milestones by a factor of gamma, set to $0.1$ in our case. Specifically, the learning rate was reduced at two stages: after $66.667$\% and $86.666$\% of the total number of training batches. And, $\lambda$ in Eq.~\ref{eq:loss} and the number of heads ($T$) in $\phi$ were set to $1$ and $8$, respectively. 

\subsection{Facial Part Segmentation}
\begin{figure}
    \centering
    \begin{subfigure}[]{0.77\textwidth}
        \centering
        \includegraphics[width=\textwidth]{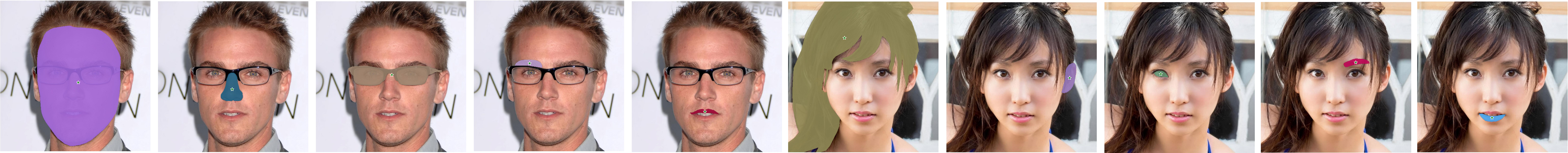}
        \caption{Ground truth}
        \label{fig:qual_face1_gt}
    \end{subfigure}
    \begin{subfigure}[]{0.77\textwidth}
        \centering
        \includegraphics[width=\textwidth]{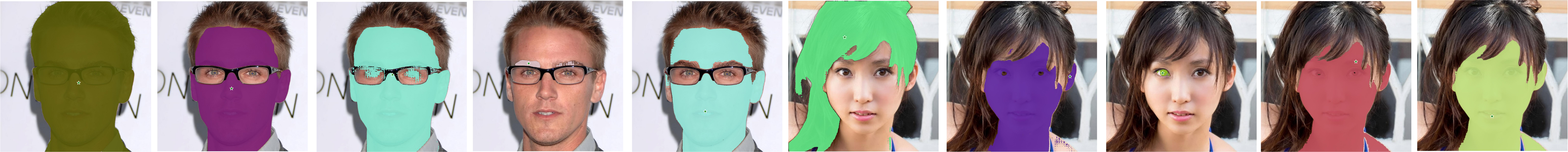}
        \caption{SAM}
    \end{subfigure}
    \begin{subfigure}[]{0.77\textwidth}
        \centering
        \includegraphics[width=\textwidth]{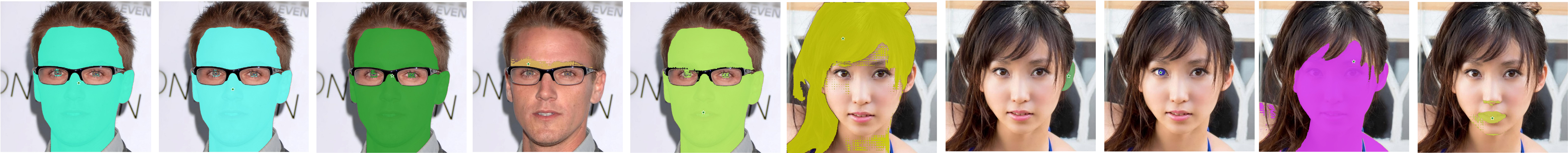}
        \caption{SAM-F}
    \end{subfigure}
    \begin{subfigure}[]{0.77\textwidth}
        \centering
        \includegraphics[width=\textwidth]{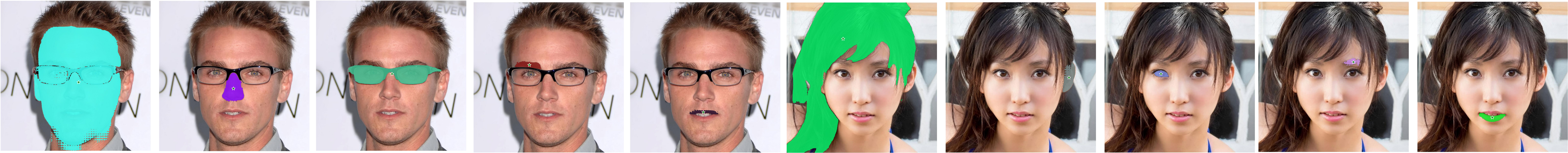}
        \caption{SAM+PLM}
    \end{subfigure}
    \begin{subfigure}[]{0.77\textwidth}
        \centering
        \includegraphics[width=\textwidth]{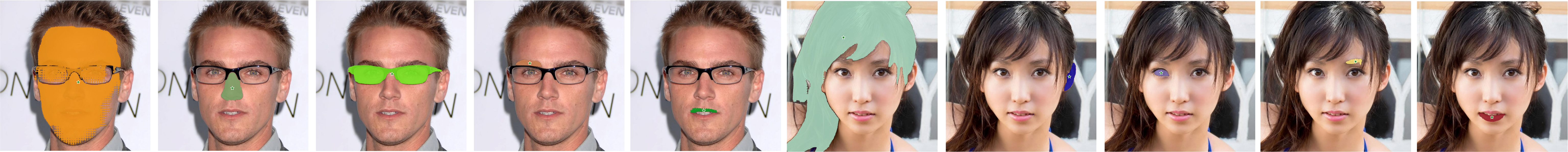}
        \caption{SAM+PLM+PMM}
    \end{subfigure}
    \vspace{-0.3cm}
    \caption{Qualitative results on facial part segmentation. Each column, sequentially from left to right, represents the following parts: skin, nose, eye glasses, right brow, upper lip, hair, left ear, right eye, left brow, and lower lip.
    }
    \label{fig:qual_face1}
    \vspace{-0.6cm}
\end{figure}

For the facial part segmentation, we utilize the CelebA-HQ dataset~\cite{CelebAMask-HQ} which provides $18$ different face part masks. 
Figure~\ref{fig:qual_face1_gt} shows sample images overlaid with GT masks for each facial part.
\begin{table*}[h!]
\small
\centering
\caption{The number of training and validation data of each part in the CelebA-HQ.}
\vspace{-0.3cm}
\resizebox{0.99\linewidth}{!}{
\begin{tabular}{l||r|r|r|r|r|r|r|r|r|r|r|r|r|r|r|r|r|r}
\Xhline{2.5\arrayrulewidth}
Part Name & `skin' & `nose' & `eye\_g' & `l\_eye' & `r\_eye' & `l\_brow' & `r\_brow' & `l\_ear' & `r\_ear' & `mouth' & `u\_lip' & `l\_lip' & `hair' & `hat' & `ear\_r' & `neck\_l' & `neck' & `cloth' \\
\Xhline{2.5\arrayrulewidth}
Training     & 21,000 & 20,999 & 979      & 20,537   & 20,531   & 20,368    & 20,310    & 10,897   & 9,903    & 12,487   & 20,913   & 20,924   & 20,551  & 880   & 5,790   & 1,307    & 20,406   & 12,009   \\ 
Validation      & 6,000  & 6,000  & 380      & 5,809    & 5,811    & 5,784     & 5,778     & 3,325    & 2,998    & 3,452    & 5,978    & 5,983    & 5,834   & 266   & 1,497   & 319      & 5,840    & 3,763    \\
\Xhline{2.5\arrayrulewidth}
\end{tabular}}
\label{table:facedataset}
\vspace{-0.3cm}
\end{table*}
Given the detailed mask for facial features from this dataset, our method can be trained to segment facial parts specifically as desired by the user. 
We used $21k$ images for training and $6k$ images for validation. Despite using identical images, the occlusions of facial parts lead to different annotation counts for each facial part. The specific counts for each part are presented in Table~\ref{table:facedataset}.


\begin{table*}[t!]
\small
\caption{Segmentation performance for facial part. It represents the mean IoU for each of the $18$ individual parts.}
\vspace{-0.6cm}
\label{tab:face}
\begin{center}
\resizebox{0.9\linewidth}{!}{
\begin{tabular}{l||ccccccccc|c}
\Xhline{2.5\arrayrulewidth}
\textbf{Method} & `skin' & `nose' & `eye\_g' & `l\_eye' & `r\_eye' & `l\_brow' & `r\_brow' & `l\_ear' & `r\_ear'  & \\
\Xhline{2.5\arrayrulewidth}
SAM & 71.37 & 12.62 & 15.57 & 50.18 & 55.83 & 24.82 & 22.71 & 44.93 & 47.97 & \\
SAM-F & 74.77 & 8.66 & 23.77 & 47.07 & 45.70 & 13.36 & 11.74 & 45.72 & 46.42\\
Proposed (SAM+PLM) & 84.38 & 82.09 & 90.10 & 76.93 & 77.57 & 61.11 & 63.76 & 76.92 & 74.37 &\\
Proposed (SAM+PLM+PMM) & \textbf{84.95} & \textbf{82.58} & \textbf{90.65} & \textbf{77.66} & \textbf{78.21} & \textbf{61.74} & \textbf{63.80} & \textbf{77.07} & \textbf{76.18} &\\
\cline{1-10}
\textcolor{gray}{SAM (\underline{oracle})} & \textcolor{gray}{80.76} & \textcolor{gray}{14.90} & \textcolor{gray}{35.88} & \textcolor{gray}{60.30} & \textcolor{gray}{59.93} & \textcolor{gray}{25.74} & \textcolor{gray}{24.30} & \textcolor{gray}{49.58} & \textcolor{gray}{50.00}  &\\
\textcolor{gray}{Proposed (SAM+PLM+PMM) (\underline{oracle})} & \textcolor{gray}{87.56} & \textcolor{gray}{83.17} & \textcolor{gray}{90.80} & \textcolor{gray}{81.22} & \textcolor{gray}{81.18} & \textcolor{gray}{64.21} & \textcolor{gray}{66.72} & \textcolor{gray}{77.11} & \textcolor{gray}{77.07}  &\\ 
\Xhline{2.5\arrayrulewidth}
\textbf{Method} & `mouth' & `u\_lip' & `l\_lip' & `hair' & `hat' & `ear\_r' & `neck\_l' & `neck' & `cloth' & \textbf{Average} \\
\Xhline{2.5\arrayrulewidth}
SAM & 55.85 & 8.57 & 20.35 & 20.35 & 58.95 & 28.62 & 29.09 & 33.91 & 29.14 & 35.05 \\
SAM-F & 30.80 & 11.03 & 21.23 & 24.12 & 64.54 & 27.51 & 31.76 & 15.24 & 29.50 & 31.83 \\
Proposed (SAM+PLM) & 76.44 & 70.80 & 73.94 & 73.54 & 82.66 & \textbf{34.77} & 54.01 & 51.47 & 76.63 & 71.19 \\
Proposed (SAM+PLM+PMM) & \textbf{76.68} & \textbf{70.88} & \textbf{74.10} & \textbf{73.88} & \textbf{83.34} & 34.17 & \textbf{55.74} & \textbf{51.59} & \textbf{76.84} & \textbf{71.67} \\ 
\cline{1-11}
\textcolor{gray}{SAM (\underline{oracle})} & \textcolor{gray}{62.75} & \textcolor{gray}{17.17} & \textcolor{gray}{31.51} & \textcolor{gray}{52.36} & \textcolor{gray}{77.29} & \textcolor{gray}{31.39} & \textcolor{gray}{36.39} & \textcolor{gray}{30.87} & \textcolor{gray}{39.44} & \textcolor{gray}{43.36} \\
\textcolor{gray}{Proposed (SAM+PLM+PMM) (\underline{oracle})} & \textcolor{gray}{78.92} & \textcolor{gray}{72.85} & \textcolor{gray}{76.69} & \textcolor{gray}{82.78} & \textcolor{gray}{86.81} & \textcolor{gray}{39.30} & \textcolor{gray}{62.11} & \textcolor{gray}{56.34} & \textcolor{gray}{79.64} & \textcolor{gray}{74.69} \\ 
\hline
\end{tabular}}
\vspace{-0.8cm}
\end{center}
\end{table*}

\textbf{Quantitative Results.} Table~\ref{tab:face} presents the single-point segmentation results for each facial part, with the `Average' indicating the mean mIoU for all $18$ parts. 
Performance is evaluated using the mask of highest confidence for both SAM and our method. For SAM, the highest IoU mask against the GT is denoted as SAM (oracle) in the table, representing an optimal selection from multiple candidate masks.
Our method, which employs prompt learning (SAM+PLM), consistently outperforms SAM, SAM-F, and SAM~(oracle). 
SAM-F, utilizing learnable weights to combine SAM's multiple masks into a final mask, often shows limited performance improvements.
This approach, while useful for blending SAM's multiple masks for objects having similar size, faces challenges in task-specific adaptation. This is due to its inability to control the multiple mask output by SAM, constraining its adaptability for customized segmentation objectives.
Moreover, incorporating the proposed PMM enhances the overall performance, except in cases of irregular shapes like earrings. 
This integration acts as an auxiliary task, serving two key roles: it prevents model overfitting on simple shapes and promotes extended training periods, thanks to its auxiliary nature enhancing point matching task.
Thus, our method confirms the effectiveness of the PMM on the generalized model, trained on large-scale datasets, to align with user intentions. Notably, even with a fixed mask decoder and solely employing prompt learning, our method outperforms SAM (oracle) in performance.

\textbf{Qualitative Results.} As shown in Fig.~\ref{fig:qual_face1}, while the original SAM performs well in larger facial areas like `skin', it struggles with accurately segmenting other facial parts such as the nose and ears. 
Despite these parts having distinct edges, their similar color and texture to the `skin' make it challenging for SAM to segment them.
SAM-F, through scale-aware fine-tuning, occasionally achieves more successful segmentation than SAM.
However, SAM-F fundamentally does not modify the mask generation process due to its reliance on SAM's mask decoder, thereby inheriting the same limitations in accurately segmenting facial parts where nuanced shape understanding is crucial. 
By contrast, our proposed method enables to adapt the segmenter in desired forms using data collected by the user. Our approach allows the creation of a SAM model that reflects the user's intent with only 0.25\% (=1.6M/641M of SAM-ViT-H) of the parameters. 

\begin{figure}[!htb]
   \begin{minipage}{0.48\textwidth}
     \centering
     \includegraphics[width=.65\linewidth]{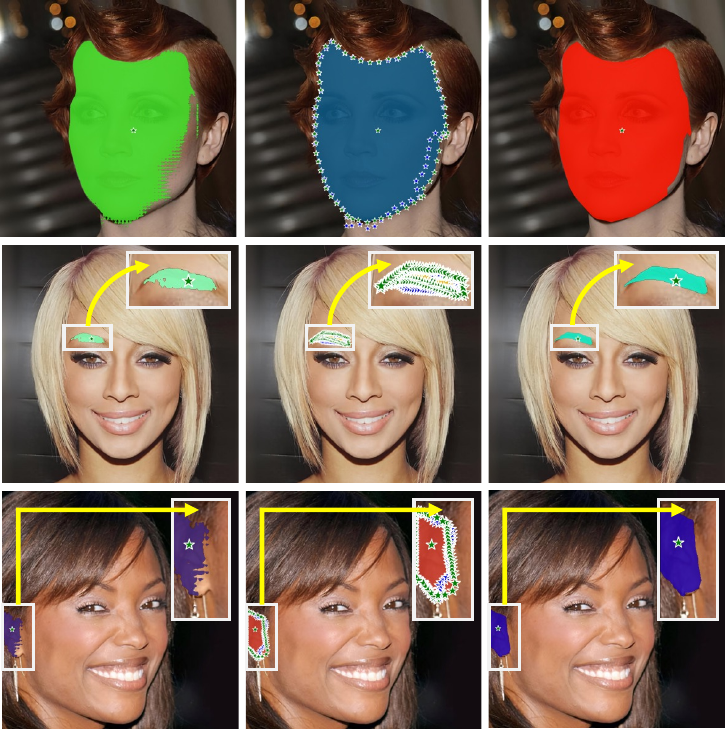}
     \vspace{-0.3cm}
     \caption{Refinement results with PMM in testing. From left to right: initial mask; point adjustments (blue dots: initial boundary, green dots: refined one); and reconstructed mask by adjusted points.}\label{fig:qual_refine}
   \end{minipage}\hfill
   \begin{minipage}{0.45\textwidth}
     \centering
     \includegraphics[width=.75\linewidth]{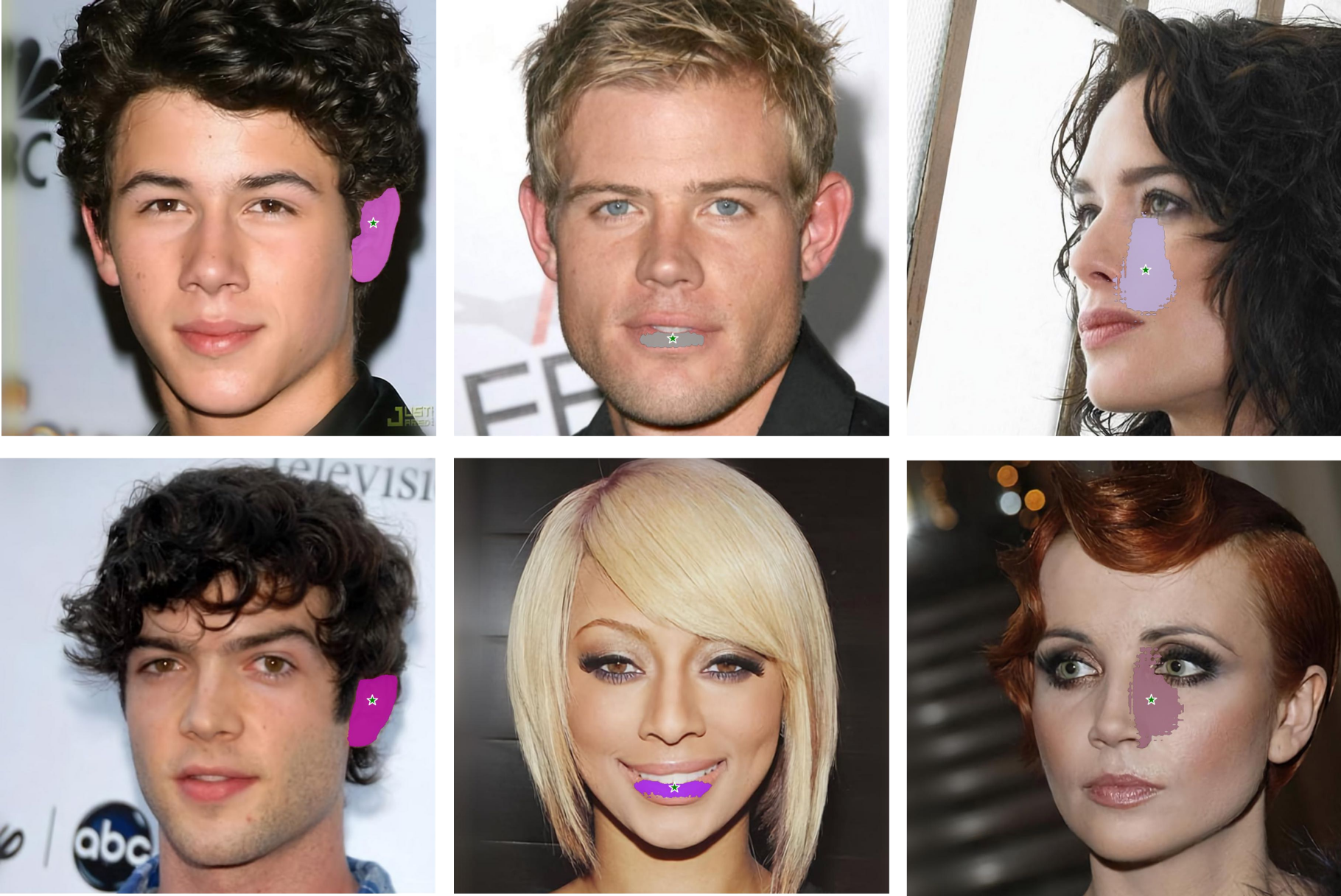}
     \vspace{0.3cm}
     
     \caption{Cross-model testing results. It shows the segmentation results when a model trained to segment a specific part is applied to a different area.}\label{fig:qual_cross}
   \end{minipage}
   \vspace{-0.3cm}
\end{figure}

\textbf{Effect of PMM.} Our method shows an additional benefit when utilizing the PMM during testing. While PMM serves as an auxiliary task in training, its application in a two-step refinement process post-training can enhance segmentation masks. Initially, we extract contour points from a mask, which are then reprocessed through the PMM to yield refined contours and a more precise polygon mask.
Figure~\ref{fig:qual_refine} illustrates these refinement results. The first column shows the initial masks from the mask decoder. The middle column visualizes the adjustment of points by the PMM, where blue dots denote the initial boundary points and green dots show the refined points. The third column shows the masks reconstructed with these refined points, leading to improved coverage and detail, especially in larger or initially vague areas like `skin'.

\textbf{Cross-model Testing.} Figure~\ref{fig:qual_cross} shows the adaptability of our method when tested across various facial parts.
The first column illustrates successful cross-application from a model trained on the right ear to segmenting the left ear, leveraging facial symmetry.
In the second column, a model trained on the upper lip is applied to the lower lip, resulting in predictions mirroring the upper lip's shape.
This indicates the distinct shape characteristics of the upper and lower lips, such as the orientations of the mouth corners. 
In the third column, a nose-trained model applied to the skin predicts a nose-shaped mask.
It shows that our approach can adapt the segmentation foundation model to produce user-specified shapes, even in the absence of clear delineations. 
This ability emphasizes the technique's effectiveness in accommodating the diverse and specific characteristics of facial features.

\subsection{Outdoor Banner Segmentation}
We applied our method to segmenting outdoor banners, a task where users typically expect rectangular segmentation outcomes.
The complexity of banners, with their text, images, and patterns, often challenges the conventional SAM, leading to imprecise segmentations.
To tackle this, we created a dataset with minimal labeling, consisting of banner and background images sourced online. Banners were randomly attached to backgrounds, with affine transformations applied to mimic different camera perspectives. The dataset featured $3k$ banner images matched with one of $4k$ background options, and we conducted tests on $980$ synthesized images using a consistent validation set. Figure~\ref{fig:qual_banner} shows that SAM usually prioritizes text or patterns, resulting in inaccurate segmentations. 
Our approach, however, enhances SAM's focus on the intended banner area, significantly improving segmentation in the presence of intricate patterns.
For a detailed comparison of performance metrics, see Table~\ref{tb:banner_license}.


\begin{table}[t]
\centering
\begin{minipage}{0.49\textwidth}
    \centering
    \caption{Segmentation performance for outdoor banner and license plate.}
    \vspace{-0.3cm}
    \label{tb:banner_license}
    \resizebox{\linewidth}{!}{
    \begin{tabular}{l||c|c}
    \Xhline{3\arrayrulewidth}
    \textbf{Method} & Outdoor Banner & License Plate \\
    \Xhline{3\arrayrulewidth}
    SAM & 30.27 & 63.79 \\
    SAM-F  & 30.60 & 53.90 \\
    Proposed (SAM+PLM) & 95.31 & 74.24 \\
    Proposed (SAM+PLM+PMM) & \textbf{97.33} & \textbf{76.29} \\\hline
    \textcolor{gray}{SAM (\underline{oracle})} & \textcolor{gray}{93.84} & \textcolor{gray}{81.98} \\
    \textcolor{gray}{Proposed (SAM+PLM+PMM) (\underline{oracle})} & \textcolor{gray}{97.44} & \textcolor{gray}{81.80} \\
    \Xhline{3\arrayrulewidth}
    \end{tabular}}
\end{minipage}
\hfill
\begin{minipage}{0.49\textwidth}
    \centering
    \caption{Number of training parameters for the SAM and the proposed modules.}
    \label{tb:params}
    \vspace{-0.3cm}
    \resizebox{\linewidth}{!}{
    \begin{tabular}{c||c}
    \Xhline{3\arrayrulewidth}
    \textbf{Module} & \textbf{Number of Parameters}\\
    \Xhline{3\arrayrulewidth}
    Image Encoder $E_{I}$ (ViT-H)  & 637M \\
    Prompt Encoder $E_{P}$ & 6.2K \\
    Mask Decoder $D$& 4.1M\\\hline
    \textbf{Prompt Learning Module (PLM) $\phi$} & \textbf{1.6M} \\
    \textbf{Point Matching Module (PMM) $\varphi$} & \textbf{1.2M} \\
    \Xhline{3\arrayrulewidth}
    \end{tabular}}
\end{minipage}
\vspace{-0.4cm}
\end{table}

\subsection{License Plate Segmentation}
As the third application of our method, we tackled license plate segmentation using the Kaggle Car License Plate dataset~\cite{kaggle_carplate}, which comprises $433$ images with bounding box annotations.
\begin{figure}[htbp]
    \begin{minipage}[b]{0.47\linewidth}
        \centering
        \begin{subfigure}[b]{\linewidth}
            \centering
            \includegraphics[height=1.0cm]{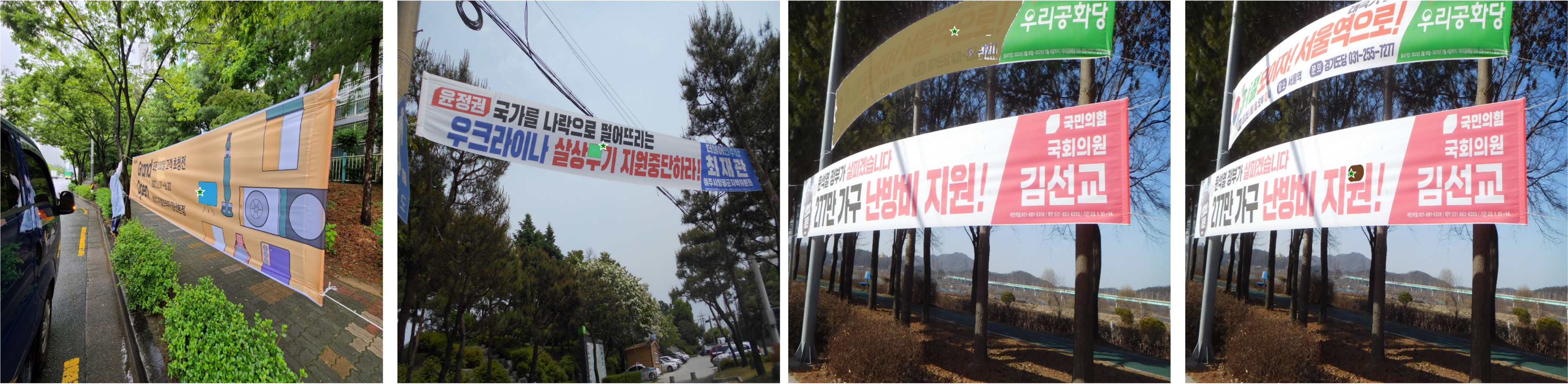}
            \caption{}
            \label{subfig:a}
        \end{subfigure}
        \vfill
        \begin{subfigure}[b]{\linewidth}
            \centering
            \includegraphics[height=1.0cm]{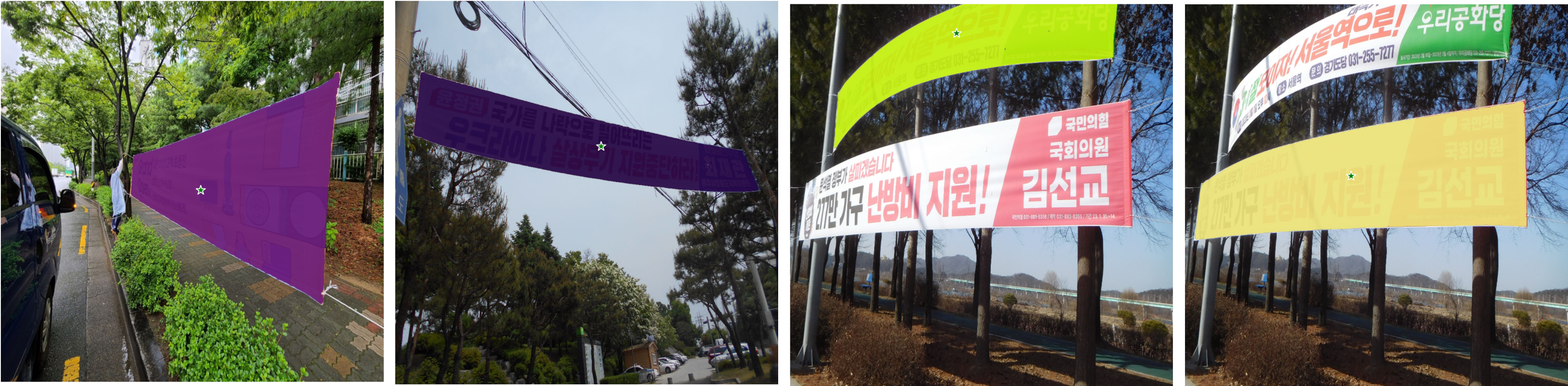}
            \caption{}
            \label{subfig:c}
        \end{subfigure}
        \vspace{-0.7cm}
        \caption{Segmentation results for outdoor banner: (a) SAM and (b) Proposed~(SAM+PLM+PMM).}
        \label{fig:qual_banner}
    \end{minipage}%
    \hfill
    \begin{minipage}[b]{0.47\linewidth}
        \centering
        \begin{subfigure}[b]{\linewidth}
            \centering
            \includegraphics[height=1.0cm]{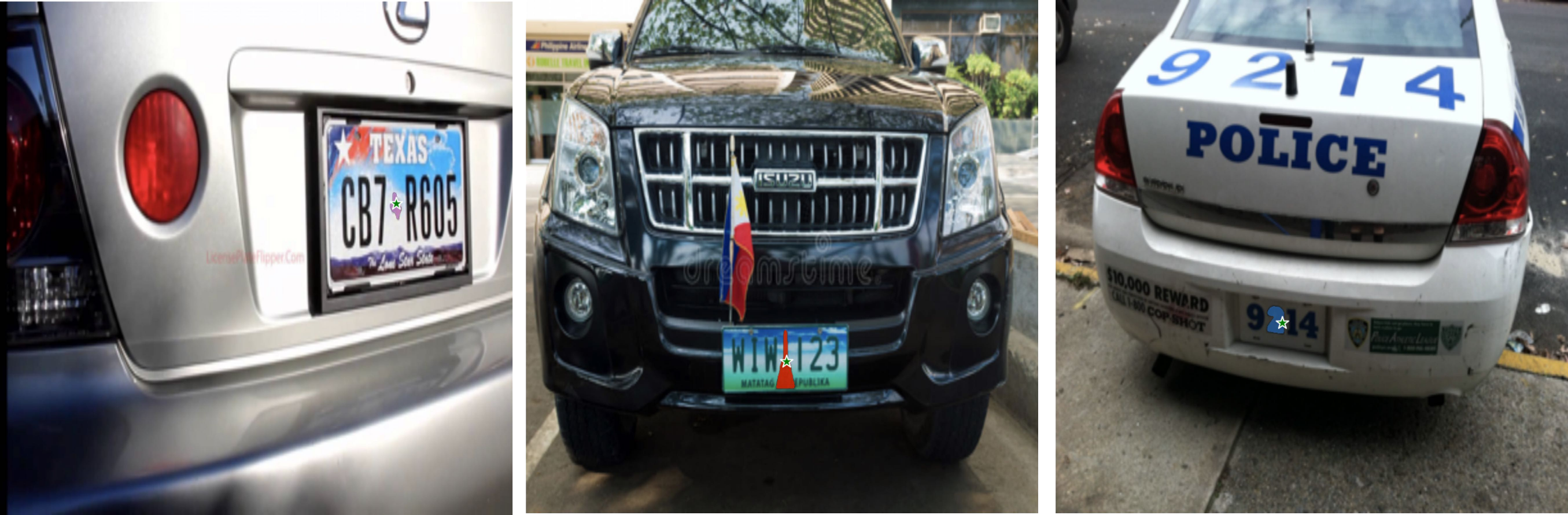}
            \caption{}
            \label{subfig:b}
        \end{subfigure}
        \vfill
        \begin{subfigure}[b]{\linewidth}
            \centering
            \includegraphics[height=1.0cm]{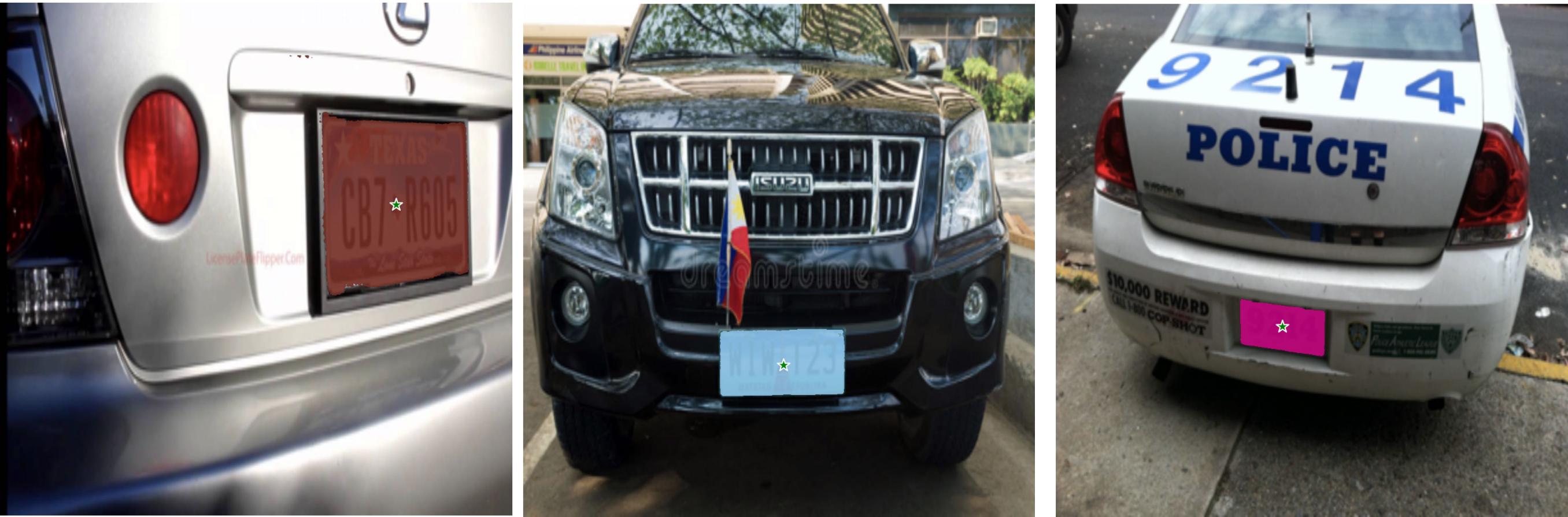}
            \caption{}
            \label{subfig:d}
        \end{subfigure}
        \vspace{-0.7cm}
        \caption{Segmentation results for license plate: (a) SAM and (b) Proposed~(SAM+PLM+PMM).}
        \label{fig:qual_license}
    \end{minipage}
\end{figure}
\begin{figure}[htbp]
    \begin{minipage}[b]{0.47\linewidth}
        \centering
        
        \vspace{-0.3cm}
        \includegraphics[height=2.2cm]{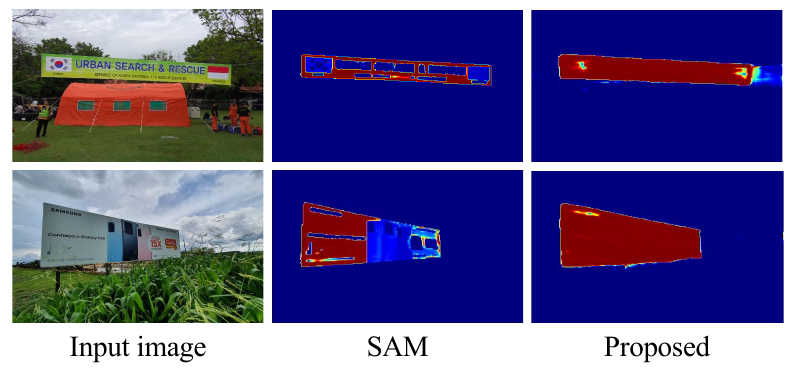}
        \vspace{-0.2cm}
\caption{Visualization of IoU maps by input prompt position.}
\label{fig:qual_robustness_test}
    \end{minipage}
    \hfill
   \begin{minipage}[b]{0.47\linewidth}
     \centering
     \vspace{-0.3cm}
    \includegraphics[height=1.7cm]{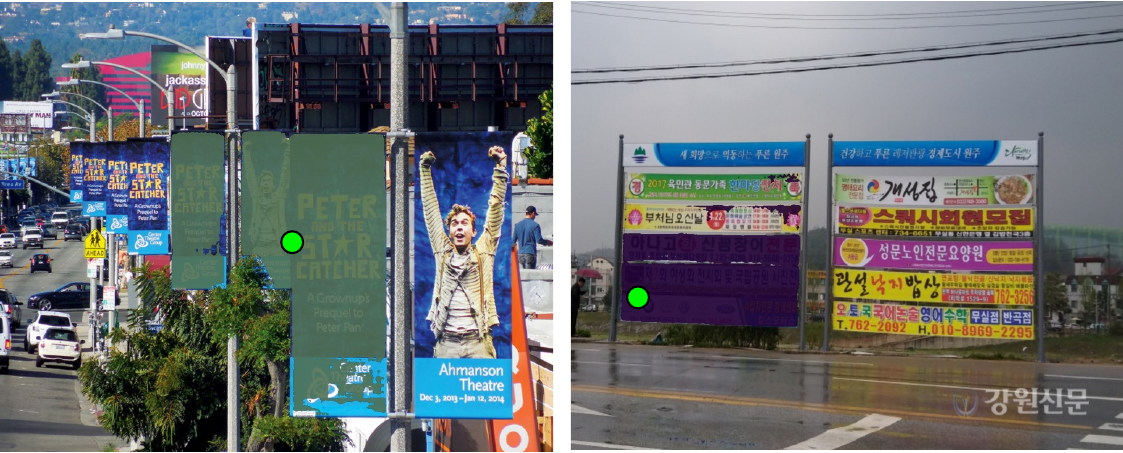}
     \vspace{0.25cm}
\caption{Example of failure cases for outdoor banner.}
\label{fig:failure_case}
   \end{minipage}
   \vspace{-0.5cm}
\end{figure}
This dataset was proportionately divided into training and validation sets with an 80:20 ratio, and annotations were converted to polygons to capture the varied shapes of license plates more accurately.
Figure~\ref{fig:qual_license} reveals that SAM typically prioritizes text or patterns, which often leads to inaccurate segmentations. 
However, with our method applied, the segmentation model is modified to infer the license plate area in alignment with user intentions. 
For a comprehensive performance analysis, refer to Table~\ref{tb:banner_license}.




\section{Discussion}
\noindent\textbf{Prompt Sensitivity.} 
To evaluate our model's robustness to the input prompt's location, we experimented with varying prompt positions and measured the IoU scores of the resulting masks against the GT. Figure~\ref{fig:qual_robustness_test} shows these findings, with the IoU map reflecting IoU scores relative to prompt positions. Our method robustly detects the banner area, accurately reflecting the user's intent in selecting positions within the banner.

\noindent\textbf{Failure Cases.} As shown in Fig.~\ref{fig:failure_case}, the proposed method may segment similar objects into one instance when they overlap or exist near the input prompt. This can be understood because the proposed method is trained that segmenting the object containing the input prompt as an instance is more important than instance-wise discrimination.

\noindent\textbf{Dataset Dependency.}
The proposed method's varying effectiveness across datasets is due to inherent differences: for banners and license plates with strong rectangular priors and clear edges, SAM performs well, limiting dramatic improvements by our method. In contrast, for less-defined shapes like facial parts, our method effectively adapts the SAM to task-specific objects, showing more significant improvements. 

\noindent\textbf{Sparse Prompt.}
For the practical applications, we initially focused on a single-point sparse prompt due to their efficiency and straightforwardness in customized segmentation tasks. As in~\cite{sam}, the proposed method can be applied with a bounding box prompt without loss of generality. However, we hypothesized that bounding boxes containing more user-defined prior information might not significantly benefit from task-specific adaptation as point prompts do.

\noindent\textbf{Number of Parameters.}
For improved task-specific adaptation, we devise an additional PLM ($\phi$) and PMM ($\varphi$) on the segmentation foundation model (\ie, SAM). 
Since only additional modules are trained on top of the SAM instead of extensive training of large model, efficient learning is possible. 
Table~\ref{tb:params} shows the number of training parameters in the SAM model (\textit{i.e.}, image encoder, mask decoder) and the proposed modules. 
With the frozen training parameters in the SAM, we train 2.8M parameters in the training stage and 1.6M parameters are used for the inference. 
Therefore, we can train the customized instance segmentation model efficiently. 
Moreover, it offers significant flexibility, allowing for plug-and-play by either excluding this module for general instance segmentation or replacing the module learned for another task.

\begin{wrapfigure}{r}{0.5\textwidth}
    \centering
    \vspace{-0.85cm}
    \includegraphics[height=1.3cm]{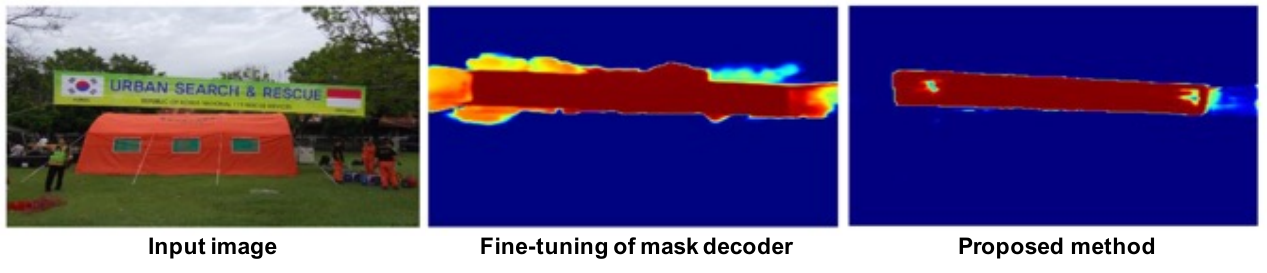}
    \vspace{-0.7cm}
    \caption{IoU maps for the proposed method and the fine-tuning of mask decoder.}
    \label{fig:decft}
\end{wrapfigure}

\noindent\textbf{Fine-tuning Mask Decoder.}
For task-specific adaptation, a straightforward approach might involve fine-tuning the mask decoder. 
Figure~\ref{fig:decft} shows IoU maps for the fine-tuning and the proposed method for outdoor banner segmentation. Compared to our method, fine-tuning the decoder with limited
 training data, which has more parameters, leads to overfitting, causing the mask region to spread when multiple objects are present. 
This shows that tuning at the encoder level is more effective than at the decoder level. 

\section{Conclusion}
In this paper, we proposed a novel task-specific adaptation to customize the segmentation foundation model via prompt learning. To tackle the issue of prompt sensitivity, we designed the prompt learning module (PLM) that transforms input prompts within the embedding space. To enhance segmentation quality, we devised the point matching module (PMM) aligning the boundary points of the estimated mask with those of the GT. 
Through experiments in customized instance segmentation scenarios, we validated the efficacy of the proposed method.
Furthermore, by training only PLM apart from the foundation model, our method can be used in a plug-and-play manner without compromising the generalization capability of the foundation model. 

\section*{Acknowledgement}
\footnotesize{This work was supported by Institute of Information \& communications Technology Planning \& Evaluation (IITP) grant funded by the Korea government (MSIT) (No.2022-0-00124, Development of Artificial Intelligence Technology for Self-Improving Competency-Aware Learning Capabilities). }

\bibliographystyle{splncs04}
\bibliography{main}

\begin{thebibliography}{10}
\providecommand{\url}[1]{\texttt{#1}}
\providecommand{\urlprefix}{URL }
\providecommand{\doi}[1]{https://doi.org/#1}

\bibitem{kaggle_carplate}
Kaggle car license plate detection.
  \url{https://www.kaggle.com/datasets/andrewmvd/car-plate-detection},
  accessed: 2024-3-7

\bibitem{layernormalization}
Ba, J.L., Kiros, J.R., Hinton, G.E.: Layer normalization. arXiv preprint
  arXiv:1607.06450  (2016)

\bibitem{medical_seg_CVPR2023}
Bai, Y., Chen, D., Li, Q., Shen, W., Wang, Y.: Bidirectional copy-paste for
  semi-supervised medical image segmentation. In: CVPR. pp. 11514--11524 (2023)

\bibitem{yolact}
Bolya, D., Zhou, C., Xiao, F., Lee, Y.J.: Yolact: Real-time instance
  segmentation. In: ICCV. pp. 9157--9166 (2019)

\bibitem{foundation}
Bommasani, R., Hudson, D.A., Adeli, E., Altman, R., Arora, S., von Arx, S.,
  Bernstein, M.S., Bohg, J., Bosselut, A., Brunskill, E., et~al.: On the
  opportunities and risks of foundation models. arXiv preprint arXiv:2108.07258
   (2021)

\bibitem{cao2023segment}
Cao, Y., Xu, X., Sun, C., Cheng, Y., Du, Z., Gao, L., Shen, W.: Segment any
  anomaly without training via hybrid prompt regularization. arXiv preprint
  arXiv:2305.10724  (2023)

\bibitem{SAM_Nerfs}
Cen, J., Zhou, Z., Fang, J., Shen, W., Xie, L., Zhang, X., Tian, Q.: Segment
  anything in 3d with nerfs. In: NeurIPS (2023)

\bibitem{SAM-OCTA}
Chen, X., Wang, C., Ning, H., Li, S.: Sam-octa: Prompting segment-anything for
  octa image segmentation. arXiv preprint arXiv:2310.07183  (2023)

\bibitem{C-cam_CVPR2022}
Chen, Z., Tian, Z., Zhu, J., Li, C., Du, S.: C-cam: Causal cam for weakly
  supervised semantic segmentation on medical image. In: CVPR. pp. 11676--11685
  (2022)

\bibitem{SAM_Track}
Cheng, Y., Li, L., Xu, Y., Li, X., Yang, Z., Wang, W., Yang, Y.: Segment and
  track anything. arXiv preprint arXiv:2305.06558  (2023)

\bibitem{solq}
Dong, B., Zeng, F., Wang, T., Zhang, X., Wei, Y.: Solq: Segmenting objects by
  learning queries. In: NeurIPS. pp. 21898--21909 (2021)

\bibitem{vit}
Dosovitskiy, A., Beyer, L., Kolesnikov, A., Weissenborn, D., Zhai, X.,
  Unterthiner, T., Dehghani, M., Minderer, M., Heigold, G., Gelly, S., et~al.:
  An image is worth 16x16 words: Transformers for image recognition at scale.
  In: ICLR (2020)

\bibitem{queryinst}
Fang, Y., Yang, S., Wang, X., Li, Y., Fang, C., Shan, Y., Feng, B., Liu, W.:
  Instances as queries. In: ICCV. pp. 6910--6919 (2021)

\bibitem{segment_for_auto_driving_IEEE_TITS2020}
Feng, D., Haase-Sch{\"u}tz, C., Rosenbaum, L., Hertlein, H., Glaeser, C., Timm,
  F., Wiesbeck, W., Dietmayer, K.: Deep multi-modal object detection and
  semantic segmentation for autonomous driving: Datasets, methods, and
  challenges. IEEE Trans. Intell. Transp. Syst.  \textbf{22}(3),  1341--1360
  (2020)

\bibitem{maskrcnn}
He, K., Gkioxari, G., Doll{\'a}r, P., Girshick, R.: Mask r-cnn. In: ICCV. pp.
  2961--2969 (2017)

\bibitem{LoRA}
Hu, E.J., Shen, Y., Wallis, P., Allen-Zhu, Z., Li, Y., Wang, S., Wang, L.,
  Chen, W.: Lora: Low-rank adaptation of large language models. arXiv preprint
  arXiv:2106.09685  (2021)

\bibitem{istr}
Hu, J., Cao, L., Lu, Y., Zhang, S., Wang, Y., Li, K., Huang, F., Shao, L., Ji,
  R.: Istr: End-to-end instance segmentation with transformers. arXiv preprint
  arXiv:2105.00637  (2021)

\bibitem{huang2019mask}
Huang, Z., Huang, L., Gong, Y., Huang, C., Wang, X.: Mask scoring r-cnn. In:
  CVPR. pp. 6409--6418 (2019)

\bibitem{align}
Jia, C., Yang, Y., Xia, Y., Chen, Y.T., Parekh, Z., Pham, H., Le, Q., Sung,
  Y.H., Li, Z., Duerig, T.: Scaling up visual and vision-language
  representation learning with noisy text supervision. In: ICML. pp. 4904--4916
  (2021)

\bibitem{VPT}
Jia, M., Tang, L., Chen, B.C., Cardie, C., Belongie, S., Hariharan, B., Lim,
  S.N.: Visual prompt tuning. In: ECCV. pp. 709--727. Springer (2022)

\bibitem{HQ-SAM}
Ke, L., Ye, M., Danelljan, M., Liu, Y., Tai, Y.W., Tang, C.K., Yu, F.: Segment
  anything in high quality. In: NeurIPS (2023)

\bibitem{sam}
Kirillov, A., Mintun, E., Ravi, N., Mao, H., Rolland, C., Gustafson, L., Xiao,
  T., Whitehead, S., Berg, A.C., Lo, W.Y., Dollar, P., Girschick, R.: {Segment
  Anything}. In: ICCV. pp. 4015--4026 (2023)

\bibitem{CelebAMask-HQ}
Lee, C.H., Liu, Z., Wu, L., Luo, P.: Maskgan: Towards diverse and interactive
  facial image manipulation. In: CVPR (2020)

\bibitem{centermask}
Lee, Y., Park, J.: Centermask: Real-time anchor-free instance segmentation. In:
  CVPR. pp. 13906--13915 (2020)

\bibitem{Power_prompt_tuning}
Lester, B., Al-Rfou, R., Constant, N.: The power of scale for
  parameter-efficient prompt tuning. In: EMNLP. pp. 3045--3059 (2021)

\bibitem{Prefix-tuning}
Li, X.L., Liang, P.: Prefix-tuning: Optimizing continuous prompts for
  generation. arXiv preprint arXiv:2101.00190  (2021)

\bibitem{coco}
Lin, T.Y., Maire, M., Belongie, S., Hays, J., Perona, P., Ramanan, D.,
  Doll{\'a}r, P., Zitnick, C.L.: Microsoft coco: Common objects in context. In:
  ECCV. pp. 740--755 (2014)

\bibitem{Editgan_Neurips2021}
Ling, H., Kreis, K., Li, D., Kim, S.W., Torralba, A., Fidler, S.: Editgan:
  High-precision semantic image editing. In: NeurIPS. pp. 16331--16345 (2021)

\bibitem{prompt_NLP}
Liu, P., Yuan, W., Fu, J., Jiang, Z., Hayashi, H., Neubig, G.: Pre-train,
  prompt, and predict: A systematic survey of prompting methods in natural
  language processing. ACM Computing Surveys  \textbf{55}(9),  1--35 (2023)

\bibitem{dino}
Liu, S., Zeng, Z., Ren, T., Li, F., Zhang, H., Yang, J., Li, C., Yang, J., Su,
  H., Zhu, J., et~al.: Grounding dino: Marrying dino with grounded pre-training
  for open-set object detection. arXiv preprint arXiv:2303.05499  (2023)

\bibitem{P-tuning_v2}
Liu, X., Ji, K., Fu, Y., Tam, W.L., Du, Z., Yang, Z., Tang, J.: P-tuning v2:
  Prompt tuning can be comparable to fine-tuning universally across scales and
  tasks. arXiv preprint arXiv:2110.07602  (2021)

\bibitem{med_SAM}
Ma, J., He, Y., Li, F., Han, L., You, C., Wang, B.: Segment anything in medical
  images. Nature Communications  \textbf{15}(1), ~654 (2024)

\bibitem{Mazurowski_et_al}
Mazurowski, M.A., Dong, H., Gu, H., Yang, J., Konz, N., Zhang, Y.: Segment
  anything model for medical image analysis: an experimental study. Medical
  Image Analysis  \textbf{89},  102918 (2023)

\bibitem{clip}
Radford, A., Kim, J.W., Hallacy, C., Ramesh, A., Goh, G., Agarwal, S., Sastry,
  G., Askell, A., Mishkin, P., Clark, J., et~al.: Learning transferable visual
  models from natural language supervision. In: ICML. pp. 8748--8763 (2021)

\bibitem{ren2015faster}
Ren, S., He, K., Girshick, R., Sun, J.: Faster r-cnn: Towards real-time object
  detection with region proposal networks. In: NeurIPS (2015)

\bibitem{segment_for_auto_driving_CVPRW2018}
Siam, M., Gamal, M., Abdel-Razek, M., Yogamani, S., Jagersand, M., Zhang, H.: A
  comparative study of real-time semantic segmentation for autonomous driving.
  In: CVPRW. pp. 587--597 (2018)

\bibitem{Tang_et_al}
Tang, L., Xiao, H., Li, B.: Can sam segment anything? when sam meets
  camouflaged object detection. arXiv preprint arXiv:2304.04709  (2023)

\bibitem{attention}
Vaswani, A., Shazeer, N., Parmar, N., Uszkoreit, J., Jones, L., Gomez, A.N.,
  Kaiser, {\L}., Polosukhin, I.: Attention is all you need. In: NeurIPS (2017)

\bibitem{SAM_remote_sensing}
Wang, D., Zhang, J., Du, B., Tao, D., Zhang, L.: Scaling-up remote sensing
  segmentation dataset with segment anything model. arXiv preprint
  arXiv:2305.02034  (2023)

\bibitem{solo}
Wang, X., Kong, T., Shen, C., Jiang, Y., Li, L.: Solo: Segmenting objects by
  locations. In: ECCV. pp. 649--665 (2020)

\bibitem{med_SAM_ada}
Wu, J., Fu, R., Fang, H., Liu, Y., Wang, Z., Xu, Y., Jin, Y., Arbel, T.:
  Medical sam adapter: Adapting segment anything model for medical image
  segmentation. arXiv preprint arXiv:2304.12620  (2023)

\bibitem{Track_anything}
Yang, J., Gao, M., Li, Z., Gao, S., Wang, F., Zheng, F.: Track anything:
  Segment anything meets videos. arXiv preprint arXiv:2304.11968  (2023)

\bibitem{Inpaint_anything}
Yu, T., Feng, R., Feng, R., Liu, J., Jin, X., Zeng, W., Chen, Z.: Inpaint
  anything: Segment anything meets image inpainting. arXiv preprint
  arXiv:2304.06790  (2023)

\bibitem{Imag_editing_SPL2020}
Zhang, J., Yang, P., Wang, W., Hong, Y., Zhang, L.: Image editing via
  segmentation guided self-attention network. IEEE Sign. Process. Letters
  \textbf{27},  1605--1609 (2020)

\bibitem{Cyclemix_CVPR2022}
Zhang, K., Zhuang, X.: Cyclemix: A holistic strategy for medical image
  segmentation from scribble supervision. In: CVPR. pp. 11656--11665 (2022)

\bibitem{PerSAM}
Zhang, R., Jiang, Z., Guo, Z., Yan, S., Pan, J., Dong, H., Gao, P., Li, H.:
  Personalize segment anything model with one shot. arXiv preprint
  arXiv:2305.03048  (2023)

\bibitem{zhang_et_al}
Zhang, S., Metaxas, D.: On the challenges and perspectives of foundation models
  for medical image analysis. arXiv preprint arXiv:2306.05705  (2023)

\bibitem{textbpn++}
Zhang, S.X., Yang, C., Zhu, X., Yin, X.C.: Arbitrary shape text detection via
  boundary transformer. IEEE TMM  (2023)

\bibitem{Zhou_et_al}
Zhou, T., Zhang, Y., Zhou, Y., Wu, Y., Gong, C.: Can sam segment polyps? arXiv
  preprint arXiv:2304.07583  (2023)

\end{thebibliography}
\end{document}